# A Step from Probabilistic Programming to Cognitive Architectures

Alexey Potapov[1,2]

[1]ITMO University, St. Petersburg, Russia
[2]St. Petersburg State University, St. Petersburg, Russia
potapov@aideus.com

**Abstract.** Probabilistic programming is considered as a framework, in which basic components of cognitive architectures can be represented in unified and elegant fashion. At the same time, necessity of adopting some component of cognitive architectures for extending capabilities of probabilistic programming languages is pointed out. In particular, implicit specification of generative models via declaration of concepts and links between them is proposed, and usefulness of declarative knowledge for achieving efficient inference is briefly discussed.

## 1 Introduction

Any AGI system should rely on some knowledge (experience) representation, learning (prediction) methods, and reasoning (action selection) methods. Although these components are not necessarily explicit, and some systems can be more syncretic than others, we can characterize approaches to AGI by them. For example, basic models of universal algorithmic intelligence like AIXI implicitly represent knowledge in the form of programs, and use Solomonoff prediction, and exhaustive search for action selection. Cognitive architectures (CA) usually utilize more restrictive representations and learning methods for the sake of computational efficiency.

Some architectures use one uniform representation and corresponding learning method yielding "grand unification and functional elegance", e.g. [1], but loosing expressiveness. Others utilize quite general knowledge representations and many inference strategies [2] that result in higher expressiveness, but causes difficulties with integrations of different components of the CA.

Achieving "grand unification and functional elegance" for more general representations can be considered as a direction of further development of CAs. Here, we claim that the probabilistic programming paradigm can be seen as a theory for CAs with the properties of grand unification and functional elegance for universal (Turing-complete) representations. We also show that insights from CAs can be very useful for further development of probabilistic programming languages (PPLs).

## 2  Basic Components of CAs in PPLs

Basic purpose of PPLs is to conduct conditional inference over generative models specified in the form of programs with random choices. One can specify models corresponding both to particular narrow machine learning methods and to a sort of universal induction (if the model generates arbitrary programs). The same inference engine can be used to solve deductive reasoning tasks (see an example with the subset sum problem in [3]). One can also perform a sort of knowledge-based reasoning using probabilistic programming for free (see, e.g. [4]).

Of course, PPLs usually don't support some distinct representation of knowledge separated from the rest code. This also has a positive side – any kind of computable knowledge can be expressed.

Thus, PPLs can be used to quite naturally and uniformly implement three basic components of AGI systems (knowledge representation, reasoning, and learning). Of course, there are some obvious differences between PPLs and cognitive architectures. PPLs only have capabilities to represent knowledge, perform reasoning and learning, which should be realized and combined. However, PPLs seem suitable as a meta-tool for designing and implementing cognitive architectures in a convenient and unified way. The real problem here is not in designing a specific architecture, but in efficiency of inference.

Indeed, inefficiency of AIXI is directly reflected in its implementation in a PPL. Turing-incomplete PPLs use more efficient inference methods, but they cannot reproduce AIXI. One possible way to try to achieve both efficiency and universality is program specialization [5]. The idea is to automatically construct an efficient projection of a universal inference method w.r.t. given specific task or generative model. If there is a program in PPL, one should not immediately apply a general inference method, but should try to optimize it w.r.t. this program.

There are some attempts to do something like this in PPLs. For example, in [6] program analysis is performed to propagate observations backward through the program. In [7] something similar to specialization of PPL inference engine w.r.t. given program is performed.

However, there is no simple and universal solution for efficient program specialization (with possibly exponential gain in speed), just like there is no simple and practically efficient universal inference method. The specializer should be an expert in program analysis, and it should be able to learn new ways to analyze and optimize programs. That is, it is impossible to put efficient and general methods inside the (static) PPL interpreter, because then such interpreter will already should be a matured AGI. Instead, the AGI system should have capabilities of becoming such an expert. Then, the question is what are the main requirements to the AGI core if they are not the efficient and general inference itself? How should we extend the paradigm of PPLs to make them more suitable both for AGI development and real-world applications?

## 3   Extending PPLs with Declarative Knowledge

Consider the following simple program in Church language [4].
```
(rejection-query
 (define x (random-integer 10))
 x
 (= (+ x 5) 10))
```
Basic PPLs will blindly search for the appropriate solution. This by itself is not necessarily bad, since if you ask a small child to find such a value that its sum with 5 is equal to 10, she or he (possessing basic knowledge about numbers) will also do this by blindly searching for the appropriate number.

More sophisticated PPLs might be able to analyze the condition, propagate it back, and infer that x is necessarily equal to 5, and this value can be sampled. One can implement complex program analysis in order to make such sampling efficient. However, it will fail in less trivial cases, in which the condition cannot be propagated backward, and non-strict heuristics or non-obvious rules should guide the search.

An AGI system should be able to solve such tasks efficiently not by some universal inference mechanism (this is impossible), but using its knowledge about numbers, arithmetic operations and equations. This is the difference between CAs and programming languages.

Also, in the context of AGI, we don't want to define the range for x unless it is known from the task. We should simply define that x is an integer. And there should be knowledge that integers can be different, and different integer values have different probabilities (in different contexts). Thus, the system should have some general generative model for integers (which is a part of its knowledge system).

Imagine that you are asked to pick a random number. How will you do this? You can pick 7, −10, 1.78324, pi, $2+3i$, etc. Apparently, humans don't use (or very rarely use as a specific case) some unbiased universal machine as a generative model for all occasions. Instead, they have declarative knowledge that numbers can be natural, integer, rational, real, complex, pi is the number, numbers consist of digits, etc.

Let us introduce a special form (is-a expression concept) that binds expression with concept via is-a link. One can think of this concept as ConceptNode and is-a link as InheritanceLink stored in Atomspace in OpenCog [2].

Concepts are similar to variables, but are bound not with values, but are linked to expressions and other concepts. Let (concept c) adds c to the concept environment of the program. Semantics of functions is extended to deal with concepts. If a concept is passed to a function instead of a variable, it is not evaluated immediately, but is considered as a symbolic expression. We can use (sample concept) to pick an arbitrary instance of this concept. Corresponding generative model is automatically constructed using multinomial distribution over all is-a links (weights of such links can be introduced and used in sampling). E.g., if one specified (is-a (normal 0 1) real-number), (is-a pi real-number), (is-a real-number number), (is-a integer number), etc., and request (sample number), then the interpreter will first randomly choose between real-number and integer, and then (if real-number is chosen) can randomly choose between pi and sampling from normal distribution.

We can also give recursive is-a definitions, e.g. (is-a null sequence), (is-a (cons number sequence) sequence). (sample sequence) will recursively construct a sequence of numbers (longer sequences will be exponentially less probably), but either number or sequence concept can be instantiated first, so we can obtain e.g. integer sequence, or some mixed sequence. Being asked 'pick a sequence of numbers', humans can also decide to use only integers. Thus, we can replace explicit priors like (define x (random-integer 10)) with priors implicitly defined by systems knowledge (define x (sample integer)). These priors can be made context-dependent with context dependent is-a weights as it is done for truth-values in OpenCog.

More principal question is how to make inference efficient using explicit knowledge. This question requires separate lengthy discussion, but the basic idea is as follows. Necessary knowledge can be specified in the form of equivalence or implication links, e.g. (equivalence (= (+ $A $B) $C) (= $A (– $C $B))) like in OpenCog. Pattern matching can be used to match (= (+ x 5) 10) against (= (+ $A $B) $C), so the entire program (not only concepts) should be put into something like AtomSpace (and this is good since programs represent knowledge in PPLs). Then, this expression should be transformed into equivalent representation (= x (– 10 5)) and simplified. The question is whether pattern matching and transformation by itself should be guided by knowledge, and if yes, what elements should be added to represent this knowledge.

## Conclusion

We considered one possible step from PPLs to CAs, and discussed its usefulness for PPLs. This doesn't mean that this way leads to yet another CA without any improvements. Benefits for CAs are "grand unification and functional elegance" without loss of generality. Adoptation of OpenCog-like pattern matching is the next step.

This work was supported by Ministry of Education and Science of the Russian Federation, and by Government of Russian Federation, Grant 074-U01.